\documentclass[letterpaper, 10 pt, conference]{ieeeconf}  % Comment this line out if you need a4paper

\IEEEoverridecommandlockouts                              % This command is only needed if 
\usepackage{multirow}
\usepackage{amsmath}  
\usepackage{amssymb}  
\usepackage[colorlinks,urlcolor=blue,linkcolor=blue,citecolor=blue]{hyperref}
\usepackage{color,array}
\usepackage{xcolor}
\usepackage{graphicx}
\usepackage{setspace}
\usepackage{CJKutf8}

\usepackage{subcaption}
\usepackage{caption}

\DeclareCaptionLabelSeparator{periodspace}{.\quad}
\usepackage{graphics} % for pdf, bitmapped graphics files
\usepackage{epsfig} % for postscript graphics files
\usepackage{times} % assumes new font selection scheme installed
\usepackage{amsmath} % assumes amsmath package installed
\usepackage{amssymb}  % assumes amsmath package installed
\usepackage[linesnumbered,ruled,vlined]{algorithm2e}

\usepackage{multirow}
\usepackage{booktabs} 
\usepackage{gensymb}  
\usepackage{bm}  
\usepackage{underscore} 
\usepackage{float}
\usepackage{amsmath} % assumes amsmath package installed
\usepackage{amssymb}  % assumes amsmath package installed

\usepackage{threeparttable}

\usepackage{bbding}
\usepackage{cite}
\usepackage{hyperref}
\hypersetup{
colorlinks  = true,
citecolor = blue,
linkcolor = blue,   
urlcolor = blue
}

\author{Jieting Zhao$^{1}$, Hanjing Ye$^{1}$, Yu Zhan$^{1}$, Hao Luan$^{2}$ and Hong Zhang$^{1*}$% <-this % stops a space
\thanks{*corresponding author (hzhang@sustech.edu.cn).}% <-this % stops a space
\thanks{$^{1}$Jieting Zhao, Hanjing Ye, Yu Zhan and Hong Zhang are with Shenzhen Key Laboratory of Robotics and Computer Vision, Southern University of Science and Technology (SUSTech), and the Department of Electronic and Electrical Engineering, SUSTech.}%
\thanks{$^{2}$Hao Luan is with the Department of Computer Science, School of Computing, National University of Singapore.}%
\thanks{This work was supported by the Shenzhen Key Laboratory of Robotics and Computer Vision under Grant ZDSYS20220330160557001.}
\thanks{Code: \url{https://github.com/zhaojieting/Part_HOE}.}%
}

\begin{document}
\begin{CJK}{UTF8}{gbsn}
\title{\LARGE \bf
Human Orientation Estimation Under Partial Observation
}
\thispagestyle{empty}
\pagestyle{empty}

\maketitle

%%%%%%%%%%%%%%%%%%%%%%%%%%%%%%%%%%%%%%%%%%%%%%%%%%%%%%%%%%%%%%%%%%%%%%%%%%%%%%%%
\begin{abstract}
Reliable Human Orientation Estimation (HOE) from a monocular image is critical for autonomous agents to understand human intention. Significant progress has been made in HOE under full observation. However, the existing methods easily make a wrong prediction under partial observation and give it an unexpectedly high confidence. To solve the above problems, this study first develops a method called Part-HOE that estimates orientation from the visible joints of a target person so that it is able to handle partial observation. Subsequently, we introduce a confidence-aware orientation estimation method, enabling more accurate orientation estimation and reasonable confidence estimation under partial observation.
% The confidence is learned by constructing an adversarial strategy between the ground truth and the predicted orientation and it can help to remove unreliable orientation estimates in downstream applications.
The effectiveness of our method is validated on both public and custom-built datasets,  and it shows great accuracy and reliability improvement in partial observation scenarios. In particular, we show in real experiments that our method can benefit the robustness and consistency of the Robot Person Following (RPF) task.
\end{abstract}

%%%%%%%%%%%%%%%%%%%%%%%%%%%%%%%%%%%%%%%%%%%%%%%%%%%%%%%%%%%%%%%%%%%%%%%%%%%%%%%%
\section{INTRODUCTION} \label{introduction}
Human orientation, which in this context specifically refers to the human yaw angle, provides crucial information for many human-robot interaction applications such as Robot Person Following (RPF)\cite{islam2019person}. RPF tasks like walking-aid robots\cite{12walkingaid}, video filming drones\cite{20drones}, and autonomous trolleys\cite{18Nikdel}, all require accurate and reliable human orientation information for the robot to calculate the desired following pose with respect to human.

% Human orientation typically requires additional computation while the 3-D human position can be directly inferred from distance measurement sensors such as LRF and RGB-D cameras.

In traditional RPF systems\cite{18Nikdel}\cite{leigh2015person}, the human orientation is assumed to be aligned with the direction of movement. Human orientation can be obtained according to human velocity direction in a global frame. However, a consistent global frame needs an additional localization algorithm, which is not necessary for RPF systems. Even if global information is provided, traditional RPF systems still fail when the human is spinning without any positional change.

In contrast, Human Orientation Estimation (HOE) using monocular images does not rely on global position information. HOE has been researched for a long time in computer vision. Most early works extract handcraft features from images and estimate orientation using machine learning. Due to the constraints of the number of data and the capacity of the machine-learning model, early works show low accuracy and reliability. The development of deep neural networks (DNN) alongside human joint detection algorithms makes it possible to accurately estimate orientation. Some methods\cite{19kpt}\cite{21monoloco} tried to leverage human joints as additional cues to the HOE task. Yu et al. \cite{19kpt} improved HOE accuracy a lot by utilizing deep networks to detect human joints and extract geometric features from the detected joints. This demonstrates that cues from human joints can play a crucial role in improving the accuracy of orientation estimation.
\begin{figure}[t]
\centering %表示居中
\includegraphics[height=8cm, width=8.5cm]
{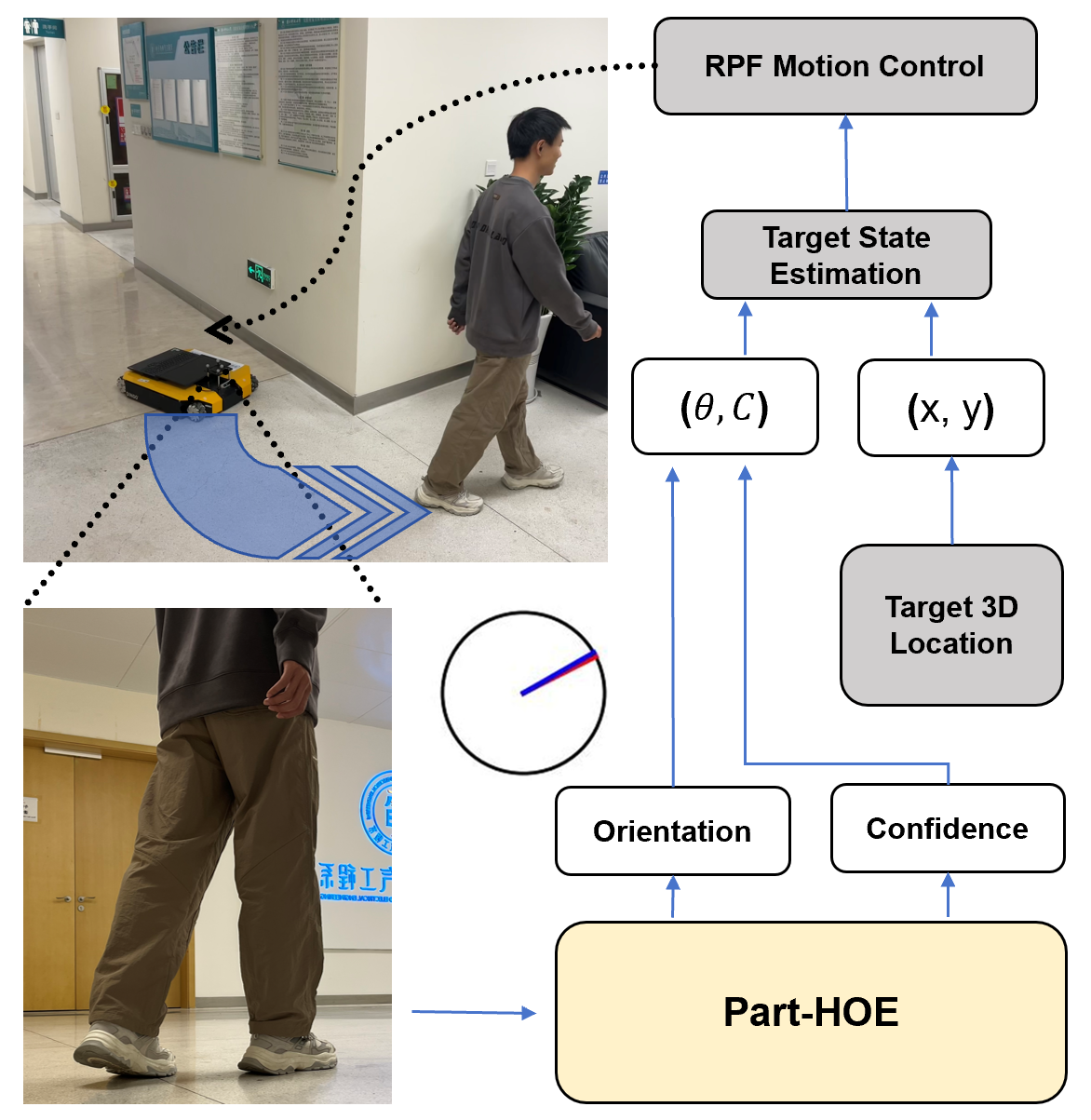}
% [height=4.5cm]表示高度
%[width=9.5cm]表示宽度
%{111.eps}表示eps格式的图片，名为111
% \captionsetup{justification=centering}
\caption{Framework Overview: An example of partial observation in robot person following. The 
existing methods for Human Orientation Estimation (HOE) struggle in this scenario. We propose an occlusion-robust method Part-HOE utilizing visible joints to help target state estimation and to improve RPF performance.}
%图片的名称
\label{Overview}
%图片的标签，用于文章中的引用，注意到标签的数字与实际文章显示的数字可能不同
\end{figure}

Despite the improvement of HOE accuracy, most orientation datasets built with a motion capture system are collected indoors with a clean background and contain no occlusion problems. As a result, algorithms developed by these datasets often struggle when confronted with partial observations, which are common in wheeled robots or robot dogs equipped with cameras. MEBOW \cite{mebow} creates in-the-wild datasets and trains orientation estimation models concurrently with human joint detection. MEBOW improves the orientation estimation accuracy a lot by learning orientation in a regression manner and uses human joint detection to provide additional cues.
Nevertheless, existing methods still struggle with partial observations because they have limited occlusion data and fail to recognize visible joints under partial observation, which are essential cues for the HOE task. In addition, MEBOW regresses the orientation with a fixed Gaussian distribution, making the predicted orientation (represented by one-hot probability distribution) not a good characteristic for filtering out low-confidence samples.

To overcome these limitations, we propose an occlusion-robust orientation estimation network by: 1) using a transformer-based network with extensive prior knowledge for joint detection.  2) 23-joint-based human body representation is used to provide additional orientation cues. Thus, our network can recognize and utilize human's visible joints to estimate the orientation. Besides, our network is able to predict reasonable confidence, which is learned
by constructing an adversarial strategy between the ground truth and the predicted orientation. 

% Furthermore, by integrating a small transformer-based human joint detection model, our approach can achieve much better accuracy with fewer parameters than the baseline network.
In summary, we propose an orientation estimation network for partial observation scenarios with confidence-aware capability. Through extensive experiments, including two public datasets and a custom-built dataset, we compare to the baseline method\cite{mebow} and yield state-of-the-art (SOTA) orientation estimation performance.  Finally, by integrating our model into an RPF system, we demonstrate our model's superiority in real RPF tasks.

\section{RELATED WORK} \label{sec:related-works}
\subsection{Human Joint Detection} \label{sec:related-works-b}
Evidence has demonstrated human joint information is helpful to the HOE task\cite{19kpt}\cite{21monoloco}\cite{mebow}. Since orientation estimation is a top-down process (estimate orientation from a cropped person image), we mainly focus on top-down human joint detection algorithms. HRNet-based algorithms\cite{hrnet}\cite{21udp}\cite{wang2023unbiased} maintain both high-resolution and deep features in multi-stages for human joint detection and achieve the SOTA accuracy on the COCO\cite{coco} dataset for about two years. MEBOW adopts the HRNet backbone with human joint detection as the auxiliary task for orientation estimation. Recently, vision transformers have shown great potential in different vision tasks, including human joint detection. Vitpose++\cite{23vitpose} adopts valina vision transformers with a transposed convolution decoder to train the human joint detection model on multiple large-scale datasets with a total of 770k samples. It surpasses the HRNet-based methods by a large margin and shows great generalization ability.

Previous HOE methods often show poor generalization problems due to the small size and low complexity of training data. Additionally, we find that HOE accuracy is highly related to the ability of human joint detection. Therefore, instead of relying on a size-limited orientation dataset, we resort to improving the human joint prediction ability of the network to improve its robustness on HOE even under partial occlusion. Specifically, we utilize a strong backbone\cite{vit} that was pre-trained on multiple large-scale joint datasets\cite{coco}\cite{mp2d}\cite{AIC}\cite{cocowhole}. Besides, to make the orientation estimation more robust when only the lower body is in the field of view, which is the most common scenario of partial observation, we adopt a human joint representation that contains human foot joints to provide more orientation cues.
\subsection{Deep Learning-based Orientation Estimation} \label{sec:related-works-a}
Deep learning is popularly used to solve the problem of orientation estimation.
Most orientation datasets are recorded indoors (due to the constraint of motion capture systems) with a simple human movement, a clean background, and full-body or upper-body observations \cite{10tud}\cite{deep-orientation}\cite{18gaze}. However, these methods are hard to generalize in the wild applications due to a lack of training samples covering observations of varied environments where partial observation is often observed.

Some research tried to use in-the-wild RGB images to train the HOE task with the help of human joint detection.
Yu et al. \cite{19kpt} utilized deep networks to detect human joints and then defined geometric features based on leg, shoulder, and hip joints. The geometric feature is then fed into an SVM model for orientation estimation. Monoloc++\cite{21monoloco} utilized a human joint detection model\cite{21pifpaf} to detect joints from in-the-wild images, and the joints' positions are then encoded to features to estimate the orientation. Due to the robustness of human joint detection, such a method improved the accuracy by a large margin. However, relying solely on joint information is not enough to provide accurate orientation estimation, particularly in cases where only a partial view of the body is available.

To further improve the accuracy of orientation estimation, MEBOW\cite{mebow} created a 72-class orientation dataset with 130k in-the-wild RGB samples. 
With large in-the-wild samples, an end-to-end HRNet\cite{hrnet} with ResNet\cite{resnet} unit architecture is adopted to train human joint detection and orientation estimation simultaneously. The orientation output is represented as a 72-class one-hot probability vector, resulting in a 5-degree resolution. Due to the similarity between adjacent orientations, MEBOW regresses the orientation probability to a fixed circular Gaussian distribution to make the network converge. With the help of MEBOW's large in-the-wild dataset and the new orientation regression strategy, the MEBOW baseline improves a lot in terms of HOE accuracy under full-body observation. 

However, MEBOW's HOE accuracy decreases a lot under partial observation due to limited occlusion
data, and the regressions strategy makes the output orientation probability hard to discriminate low-confidence samples. We solve these problems by harnessing the extensive prior knowledge from the human joint detection model. Additionally, inspired by out-of-distribution (OOD) research\cite{learnconfidence}, we propose to learn reliable confidence prediction from training data without explicit confidence labels.

\begin{figure*}
\centering 
\includegraphics[width=0.95\linewidth]{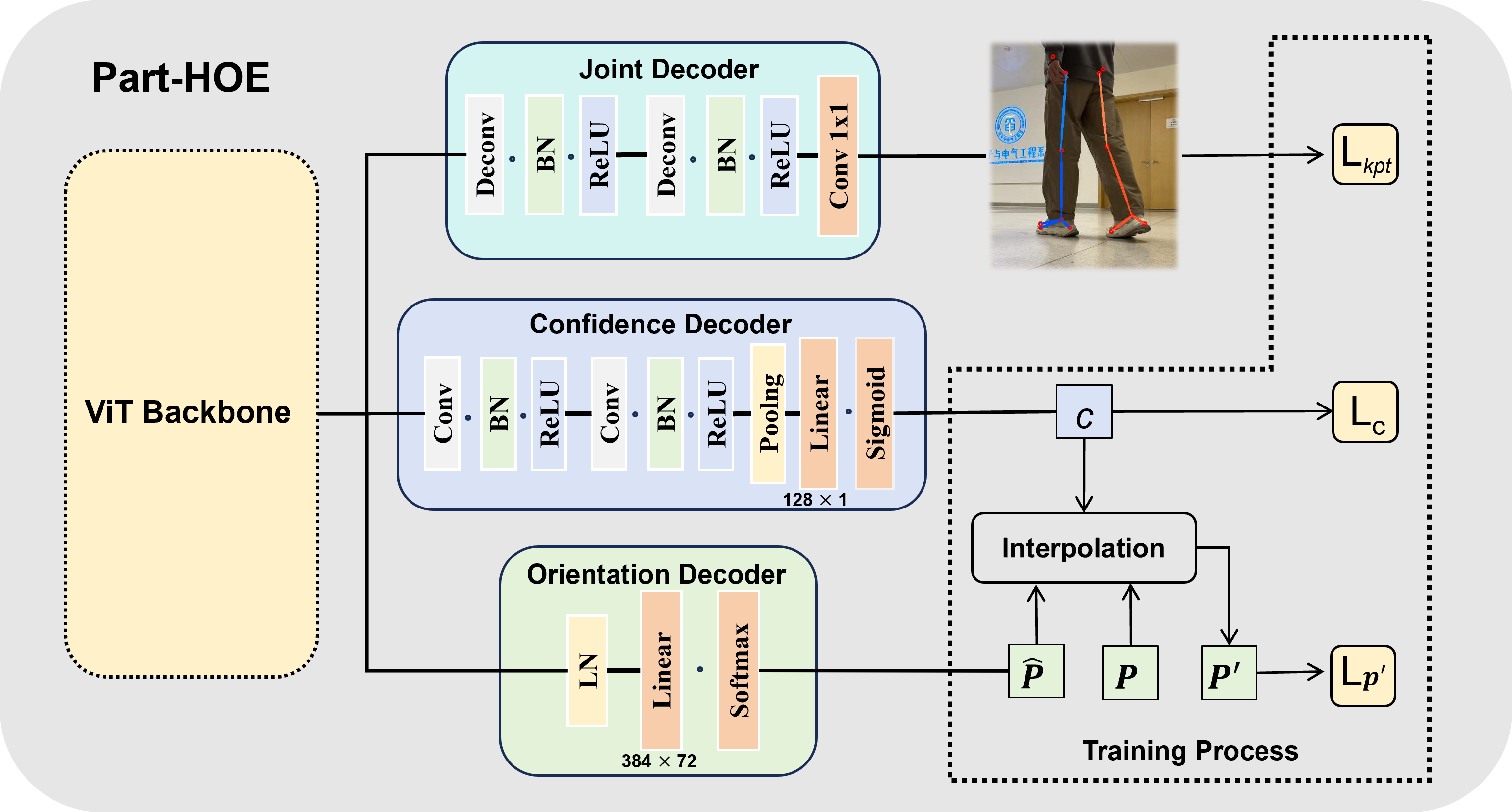}
\caption{Our Part-HOE method takes an RGB image as input and extracts features through the ViT backbone. Then, three decoder modules output orientation estimation and confidence estimation, along with 2D human joint detection. Finally, the network is learned by multi-task training.}
\label{framework}
\vspace*{-0.1in}
\end{figure*}
\subsection{Motion-based Orientation Estimation in RPF Systems} \label{sec:related-works-a}
% The most simple RPF system can be achieved in RGB camera where the robot just needs to keep the target person in the field of view of the sensor with a proportional controller.
% In the early research of the RPF system \cite{marchetti2012autonomous}, where the RPF system performs a distance serve strategy, only the following distance is considered important. 

For intelligent RPF applications such as walking-aid robots\cite{12walkingaid}\cite{08walkingaid}\cite{22walkingaid} and autonomous shopping carts\cite{18Nikdel}, HOE is necessary since robots need to maintain a desired relative pose to the target person rather than merely keeping a fixed distance. Despite the use of various sensors across different applications, typical RPF systems incorporate target state estimation and robot motion control, as shown in Fig.~\ref{Overview}. With different sensors, the robot can detect and track a person on the ground plane and further get the state estimation of the target person. 

P. Nikdel et al. \cite{18Nikdel} proposed an autonomous cart that maintains to be in front of a moving human, and the autonomous cart tracks the human's position in the ground plane with a constant motion model. The human orientation is estimated by calculating the direction of the velocity, maintaining observation of $(x, y, vt, \theta)$, where orientation $\theta = \arctan\left(\frac{y_t - y_{t-1}}{x_t - x_{t-1}}\right)$.
However, such a method suffers when the target person is static or turns with a small position change. Besides, it needs global position information which is obtained from an additional localization algorithm. Qingyang et al.\cite{22walkingaid} tried to solve the first problem by defining three different motion modes. For example, the orientation of a static human is the perpendicular vector between the left leg and the right leg. However, the state of the human lower limb can be very different for each individual, for example, people can stand by crossing legs. This method is hard to fit the general scenario.
Instead of relying on a global frame or leg-specific information, we estimate the human's orientation directly from an image. Our estimation is confidence-aware and reliable even under partial observation. We show that when the RPF system is equipped with our Part-HOE, the following behavior is more consistent than the traditional RPF system in situations of backward and forward following.

% A. Leigh et al. \cite{leigh2015person}used a laser range finder to detect t the position of a person's feet, considering the pedestrian's position as the observation and the pedestrian's velocity as the state, with the direction of the pedestrian's velocity being considered as the pedestrian's orientation. 

\section{METHODOLOGY} \label{sec:methodology}

\subsection{Overview and Problem Statement}
To estimate a person's orientation under partial occlusion, we propose a novel orientation estimation method (\textbf{Part-HOE}) considering both accuracy and confidence, as shown in Fig.~\ref{framework}.
Given a standardized RGB image of a human, we utilize ViT backbone\cite{vit}\cite{23vitpose} to extract features for the reason that the ViT is pre-trained on multiple large human joint detection datasets and has extensive human joint cues for orientation estimation. The extracted features are then fed into the joint decoder, orientation decoder, and confidence decoder.

The joint decoder uses a simple transpose convolution for up-sampling, followed by a $1\times1$ convolution layer that predicts the 23 joint heatmaps, including the foot joints (Sec.~\ref{jointdetection}). 
% The loss of the joint detection task is given the highest weight, as we found that improved accuracy of joint detection directly enhances the precision of orientation estimation.
% Since we adopt the backbone pre-trained on large-scale joint detection data, the joint detection shows a great result. helps to improve the accuracy of orientation estimation, which has been proved in previous works\cite{19kpt}\cite{mebow}\cite{21monoloco}.
The orientation decoder is an extremely simple network, which (Sec.~\ref{orientationestimation}) is composed of only a normalization layer and a fully connected layer connected to a softmax operation. When joint detection is accurate enough, we found that increasing the complexity of the orientation decoder shows a small benefit to the accuracy improvement.
The last decoder is the confidence decoder. By extracting features from the ViT output using convolution, with linear layer and sigmoid operation, the confidence output ranges from 0 to 1. Since there is no explicit confidence label, the confidence estimation is learned in a self-supervised manner as described in Sec.~\ref{confidenceestimation}. Finally, we integrate our proposed HOE method into an  RPF system (Sec.~\ref{RPFsystem}).

\subsection{Part-HOE Model}
\subsubsection{Auxiliary Human Joint Detection}
\label{jointdetection}
We do not directly use the output of joint detection for orientation estimation; instead, we use the feature map as the input to the orientation decoder to avoid losing other information. The joint detection here is trained as an auxiliary task. Here, we make an effort to enhance the HOE ability by two tricks: 1) applying a ViT-Small backbone with extensive prior knowledge of human joint detection and 2) adding more human joint constraints for the model to get additional cues from human joints.
Transformer-based algorithm ViTPose++\cite{23vitpose} achieved the state-of-the-art (SOTA) human joint detection performance on the COCO\cite{coco} dataset. By pre-training on ImageNet with MAE task and conducting human joint detection training on multiple datasets, ViTPose++ shows great generalization ability and accuracy on different datasets. To harness the prior knowledge in the human joint detection area, we use the vision transformer backbone, which is pre-trained on the joint dataset with 770K samples\cite{coco}\cite{mp2d}\cite{AIC}\cite{cocowhole}. 

Baseline MEBOW\cite{mebow} shows SOTA HOE accuracy under full-body observation. However, the accuracy decreases a lot under partial observation, especially when only lower bodies are observed. To improve the robustness and accuracy under partial observation, especially lower-body observation, we add the additional six-foot joints to the original 17-joint human representation in the COCO\cite{coco} dataset to our 23-joint-based human representation. 

The loss function for supervising human joint heatmap can be defined as the mean squared error (MSE) between the predicted heat map $\hat{H}$ and the ground truth heat map $H$. For $N$ human joints and an image with $W \times H$ pixels, the loss function is given by:
\begin{equation}
\label{eq2}
\mathcal{L}_{kpt} = \frac{1}{N} \sum_{i=1}^{N} \frac{1}{W \times H} \sum_{x=1}^{W} \sum_{y=1}^{H} \left( \hat{H}_{i}(x, y) - H_{i}(x, y) \right)^2
\end{equation}
where $\hat{H}_{i}(x, y)$ is the predicted intensity of the heatmap at pixel location $(x, y)$ for human joint $i$, and $H_{i}(x, y)$ is the corresponding ground truth intensity.

\subsubsection{Orientation Estimation}
\label{orientationestimation}
Given ViT feature map, the orientation decoder estimates the orientation as a discrete formula same as the baseline\cite{mebow}, denoted as $\hat{\textbf{p}}=[\hat{p_0},\hat{p_1},\hdots,\hat{p_{71}}]$  $(\sum_{i=0}^{71} \hat{p_i} = 1)$, where the max $\hat{p_{i}}$ indicates the orientation of a person is within the range of $[i \cdot 5^\circ-2.5^\circ, i \cdot 5^\circ+2.5^\circ]$. Here, the orientation in a range of $[0^\circ, 360^\circ)$ follows the same definition as MEBOW\cite{mebow}.

% The cross-entropy loss for the traditional classification method failed to converge in the orientation estimation task. 
We found that when joint detection is accurate enough, 
increasing the complexity of the orientation decoder yields no benefit. Therefore, our orientation decoder is composed of only a normalization layer with a fully connected layer connected with softmax operation as shown in the orientation decoder in Fig. \ref{framework}.

\begin{figure}[H]
\centering 
\includegraphics[width=0.60\linewidth]{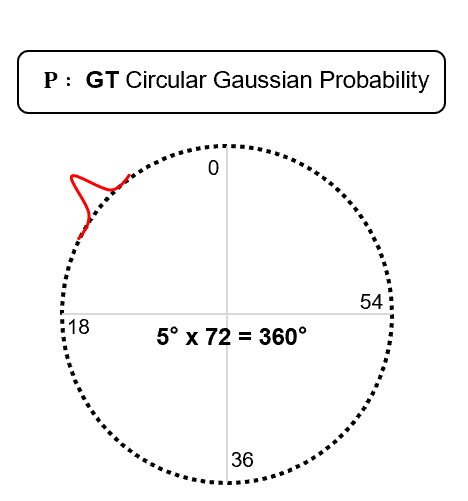}
\caption{An explanation of Circular Gaussian Probability in the interpolation operation of Part-HOE.}
\label{gaussian}
% \vspace*{-0.1in}
\end{figure}

We converted the orientation labels $l_{\text{gt}} \in [0, 71] \cap \mathbb{Z}$  to “circular” Gaussian probability $\textbf{p}$ as MEBOW\cite{mebow}, $\textbf{p} = [p_0,p_1,\hdots,p_{71}](\sum_{i=0}^{71} p_i = 1)$ and trained it as a regression task:
 \begin{equation} \label{eq4} p_{i} = \frac{1}{\sqrt{2\pi}\sigma} e^{-\frac{1}{2\sigma^2} \min(i - l_{gt}, 72 - i - l_{gt})^2}
\end{equation}
where $\sigma$ is a constant value, and the “circular” Gaussian probability is visualized in Fig. \ref{gaussian}.

\subsubsection{Confidence Estimation}
\label{confidenceestimation}
The confidence decoder outputs a confidence value range from zero to one, indicating if the orientation estimation is reliable. As we mentioned in Sec.~\ref{orientationestimation}, the training of orientation is a regression task since the loss could not converge if using traditional classification loss, i.e. cross-entropy loss. Therefore, the model regresses orientation probability to constant Gaussian distribution (Fig.~\ref{gaussian}), which means the output $\hat{\textbf{p}}$ is a regressed distribution instead of a probability. In partial observation scenarios, the model tends to predict orientations with the same distribution even when the prediction is highly unreliable. Inspired by\cite{learnconfidence}, we proposed a confidence-aware orientation estimation method to predict the confidence of each orientation estimation as shown in Fig.~\ref{framework}.  
Even without the annotation of confidence that is available during training, confidence can be learned by constructing an adversarial strategy between the ground truth and the predicted orientation. Denote the predicted confidence $c \in (0,1)$, and construct $\textbf{p}'$ for training confidence $c$ and orientation $\textbf{p}$ at the same time:
 \begin{equation} \label{eq4} \textbf{p}' =  c \cdot \hat{\textbf{p}} + (1-c) \cdot \textbf{p}\\
 \end{equation}
The loss function of the constructed $\textbf{p}'$ is defined as:
 \begin{equation} \label{eq5}\mathcal{L_{\textbf{p}'}} = \sum_{i=0}^{71} (p'_i - p_i)^2
 \end{equation}
A penalty $\mathcal{L}_{c}$ for confidence $c$ along with ${L_{\textbf{p}'}}$ for preventing confidence $c$ approaching 0. The penalty loss is expressed as:

 \begin{equation} \label{eq6} \mathcal{L}_{c} = -log(c)
 \end{equation}
 If the observation is reliable, the confidence $c$ will converge towards one; conversely, confidence $c$ will approach zero.

The final loss function is the sum of the orientation loss, the penalty confidence loss, and the joint detection loss. 
Here, $\lambda$ is a weight coefficient that dynamically changes with $\mathcal{L}_{c}$, used to balance the loss value between orientation and confidence.
 \begin{equation} \label{eq7} \mathcal{L} = \mathcal{L}_{\textbf{p}'} -\lambda \cdot \mathcal{L}_{c} + \mathcal{L}_{kpt}\\
 \end{equation}
  
% If $\mathcal{L}_{c}$  bigger than a threshold,    $\lambda = \frac{\lambda}{0.99}$, else $\lambda = 0.99\lambda$.
    
\begin{table*}[ht]
    \centering
    % \captionsetup{justification=centering}
    \caption{Evaluation of orientation accuracy on three datasets.
    Acc (N°) is the estimation accuracy that estimation is regarded as true if the estimated orientation error is within (-N°, +N°). MAE (°) is the mean absolute error of orientation. The data inside the bracket is evaluated under full observation, while those outside the bracket are evaluated under partial observation.
    }
    \label{tab1}
    \scalebox{0.9}{\begin{tabular}{c|c|cccccc}
    \toprule
        \textbf{Dataset} & \textbf{Methods}  & \textbf{GFlops} $\downarrow$ & \textbf{Params} $\downarrow$ & \textbf{Acc (5°)} $\uparrow$ & \textbf{Acc (15°)} $\uparrow$ & \textbf{Acc (30°)} $\uparrow$ & \textbf{MAE (°)} $\downarrow$ \\ \midrule
    
            MEBOW& MEBOW & 8.28G & 39.6M & 47.0\% (68.6\%) & 76.6\% (90.7\%) & 90.0\% (96.9\%) & 16.3 (8.4) \\
            
            % & Unbiased-MEBOW  & 9.61G & 30.8M & 48.5\% (\textbf{\underline{69.8\%}}) & 77.6\% (91.1\%) & 90.8\% (96.9\%) & 16.1 (8.2) \\
        
            % & Baseline & 128 × 96 & 2.07G & 39.6M & 65.2\%(46.7\%) & 88.8\%(76.2\%) & 95.6\%(89.2\%) & 9.9(17.4) \\
            & Monoloco++  & 8.70G & 25.5M & 42.6\% (49.8\%) & 75.0\% (81.5\%) & 91.3\% (93.5\%) & 14.6 (12.5) \\
            % & Ours & 128 × 96 & 2.63G & 30.8M & 67.8\%(46.9\%) & 89.9\%(75.2\%) & 96.3\%(89.2\%) & 9.1(17.2) \\
    
            & Ours & \textbf{\underline{5.62G}} & \textbf{\underline{24.2M}} & 
            \textbf{\underline{52.4\% }} (\textbf{\underline{70.1\%}})  & \textbf{\underline{82.0\%}} (\underline{\textbf{91.2\%}}) &
            \textbf{\underline{93.4\%}} 
            (\textbf{\underline{96.9\%}}) & 
            \textbf{\underline{13.4}}
            (\textbf{\underline{8.1}})
            \\ 
            
            % & Ours-B  & 8.22G & 89.9M & 53.4\% (69.8\%) & 83.4\% \textbf{(\underline{91.8\%})} & \textbf{\underline{93.7\%} (\underline{97.3\%})} & \textbf{\underline{13.4} (\underline{7.9})} \\ 
    \midrule
            H3.6M & MEBOW  & 8.28G & 39.6M & 11.7\% (30.8\%) & 31.7\% (\textbf{\underline{77.8\%}}) & 55.6\% (98.3\%) & 43.8 (10.0) \\
            & Ours & \textbf{\underline{5.62G}} & \textbf{\underline{24.2M}} & \textbf{\underline{18.7\%} (\underline{32.4\%})} & \textbf{\underline{49.8\%}} (77.1\%) &  \underline{\textbf{77.8\%}} (98.3\%) & \textbf{\underline{21.7} (\underline{9.9})} \\ 
    \midrule
            Custom-built & MEBOW  & 8.28G & 39.6M & 12.0\% & 37.1\%  & 66.8\% & 34.7 \\
            & Ours & \textbf{\underline{5.62G}} & \textbf{\underline{24.2M}} & \textbf{\underline{17.7\%}} & \textbf{\underline{47.8\%}} &  \underline{\textbf{83.4\%}} & \textbf{\underline{21.0}} \\ 
        \bottomrule
    \end{tabular}}
    % \vspace*{-0.1in}
    % \vspace*{10in}
\end{table*}
 
\subsection{Robot Person Following System}
\label{RPFsystem}
We implement an RPF system that can follow the target person both forward and backward to demonstrate the importance of Part-HOE in real RPF applications. 
Following forward or backward requires the robot to maintain a fixed distance in the front or back of the target person. The location of the target person is estimated using a leg tracker\cite{leigh2015person}. Denote the pose of the target person as $(x_{ped}, y_{ped}, \theta_{ped})$ in the robot's local frame. The ideal following backward pose $(x_{backward}, y_{backward}, \theta_{backward})$ is calculated by：
\begin{equation}
\begin{aligned}
x_{backward} &= x_{ped} - \cos(\theta_{ped}) \\
y_{backward} &= y_{ped} - \sin(\theta_{ped}) \\
\theta_{backward} &= \theta_{ped}
\end{aligned}
\label{eq7}
\end{equation}
The ideal following forward pose is calculated by：
\begin{equation}
\begin{aligned}
x_{forward} &= x_{ped} + \cos(\theta_{ped}) \\
y_{forward} &= y_{ped} + \sin(\theta_{ped}) \\
\theta_{forward} &= \theta_{ped} + \pi\\
\end{aligned}
\label{eq8}
\end{equation}
Both following forward and backward tasks need to estimate the position and orientation of the target person accurately. The RPF system includes a state estimation module and robot control module as shown in Fig.~\ref{Overview}. The details of other modules can be found in our previous work\cite{ye2023icra}\cite{ye2023person}.
Here, we integrated our Part-HOE network into the state estimation module of the RPF system.

\section{EXPERIMENTS} \label{sec:experiments}
To validate the accuracy and robustness of our proposed Part-HOE model, we conducted comparisons on three different orientation datasets, and we also evaluated it in RPF tasks by integrating our method into a robot system.
In this section, we first introduce the datasets, the baselines, and the details of the implementation of our experiments. Secondly, we demonstrate the superiority of our Part-HOE in terms of HOE accuracy and discrimination ability for low-confidence samples. An ablation study is conducted to verify the importance of different modules. Lastly, we show that Part-HOE can improve the robustness and consistency of RPF tasks.

\subsection{Datasets}
\label{datasets}
In this work, three datasets are used in the experiment, including two public datasets (MEBOW\cite{mebow} dataset and Human-3.6M\cite{human36m}) and our custom-built dataset.
\begin{itemize}
    \item The MEBOW dataset contains 130k in-the-wild samples, and the orientation is annotated to 72 class labels. The label i indicates the orientation of the person is within the range of $[i \cdot 5^\circ-2.5^\circ,  i \cdot 5^\circ+2.5^\circ]$, resulting in a 5-degree resolution.
    \item The Human3.6 M dataset provides 3D human joint annotations for continuous monocular image sequences containing different actions. To validate our model in the continuous image sequence, we also evaluated our model on the walking sequences with 158K full-body observation samples in total, where the orientations range from $[0^\circ, 360^\circ)$ can be calculated by following the definition of MEBOW\cite{mebow}.
    \item Our custom-built dataset includes 5k partial observation images recorded under a motion capture system with a helmet and a robot equipped with an RGB-D camera and motion capture markers. The orientation annotation is continuously calculated by the transformation between the helmet and the robot, and it is recorded while the robot performs a real RPF task.
\end{itemize}

\subsection{Baseline Methods}
For comparison in terms of orientation estimation accuracy, we compared with the strong baseline MEBOW  and Monoloco++. We also conducted multiple comparisons of confidence and improvement in RPF tasks with the current SOTA MEBOW.
\begin{itemize}
\item MEBOW\cite{mebow}: We compare the accuracy of the MEBOW baseline method on its validation dataset and the Human3.6M Walking dataset. To get the partial observations, we cropped only the image lower than the hip joint for evaluation.
\item Monoloco++\cite{21monoloco}: The Monoloco models take 2d human joint as input and output continuous orientation in a range of $[-\pi, \pi]$. To make a fair comparison, we retrain the monoloco network on the MEBOW dataset and provide input with ground truth human joint.
% \item Unbiased HRNet\cite{wang2023unbiased}: To make a comparison of orientation head and human joint detection accuracy, we reduced the decoder of MEBOW baseline from 3 ResNet units to 1 unit, and added additional foot joint constraints and accuracy utilizing unbiased feature alignment techniques.  We can make a new model with the same HRNet backbone as MEBOW and a much smaller header for the orientation decoder to prove that the orientation decoder is not important.
\end{itemize}
We also make a comparison with the traditional RPF algorithm\cite{22walkingaid}\cite{18Nikdel}\cite{leigh2015person}, where the state estimation of humans is tracked using the constant velocity model. To exclude the influence of position estimation, we directly obtain the ground truth of the human's positions from a motion capture system in 10 Hz.

% \subsection{Baselines}
% To validate the accuracy of the orientation estimation algorithm, we compared it with MEBOW, Monoloc, and 3D pose estimation algorithms in terms of orientation estimation accuracy. For evaluating the confidence of orientation estimation, we compared it with SOTA Baseline.
\subsection{Implementation Details}
The implementation of our model is based on ViTpose++\cite{23vitpose} and MEBOW\cite{mebow}, and the backbone was pre-trained on the human joint detection task. ViT-Small is used as our backbone, considering the computation cost. For the MEBOW dataset and Human3.6M, inputs were processed by cropping the target person using the ground truth bounding box, followed by a standardized operation. Our custom-built dataset employed YOLOX\cite{ge2021yolox} for bounding box detection, subsequently applying the same standardization procedure. Our model is trained on the MEBOW dataset \cite{mebow}, where augmentation techniques, including cropping, flipping, and scaling, are used during the training process. We use Adamw optimizer (learning rate is equal to 0.001) to train our Part-HOE for 80 epochs.

For RPF implementation, a Clearpath Dingo-O and a
laptop with Intel(R) Core(TM) i5-10200H CPU @ 2.40GHz
and NVIDIA GeForce GTX 1650 are used. A Realsense
D435i with 1280 × 720 resolution and 30 Hz frequency is
mounted on the robot with a $30^\circ$ tilt angle and a height of 0.17m relative to the ground plane. To exclude the influence of other modules, we utilize a motion capture system to obtain the human's positions, and the orientation is estimated by the baseline MEBOW or our Part-HOE. For the traditional RPF implementation, we follow the code in \cite{leigh2015person}. The control algorithm employs model predictive control, executing the robot person following both backward and forward. (see Sec.~\ref{RPFsystem})

\begin{figure}[t]
\centering %表示居中
\includegraphics[width=0.90\linewidth]{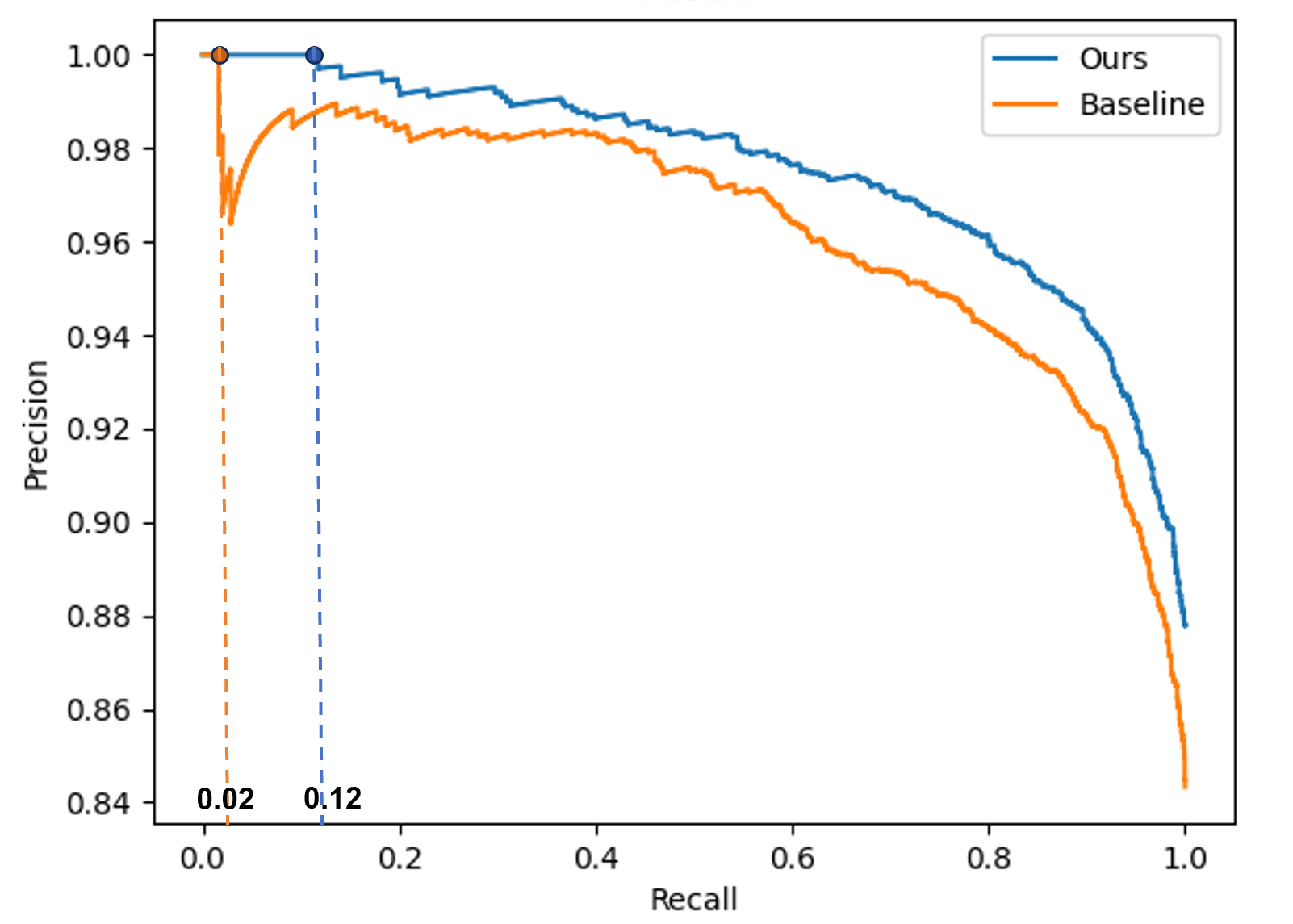}
% [height=4.5cm]表示高度
%[width=9.5cm]表示宽度
%{111.eps}表示eps格式的图片，名为111
\caption{Precision-recall curve under partial observation. The dashed lines indicate the max recall at 100\% precision.}
%图片的名称
\label{figure4}
%图片的标签，用于文章中的引用，注意到标签的数字与实际文章显示的数字可能不同
% \vspace*{-0.2in}
\end{figure}

\subsection{Experimental Results}
\subsubsection{Evaluation of Orientation Accuracy}
\label{accuracy}
The evaluation of orientation accuracy is conducted on three datasets: the MEBOW dataset, the Human 3.6M Dataset, and our custom-built dataset. Two metrics are used to evaluate the orientation accuracy, including the percentage of error within $n^\circ$ and the mean absolute error (MAE). We also evaluate the model computation cost, considering it is used on real robot tasks. In terms of computational cost, our model achieves 32\% fewer Flops and 39\%  fewer parameters compared to the baseline MEBOW as shown in Table \ref{tab1}. In terms of orientation estimation accuracy, we evaluate our method on both full-body and partial-body observation, and our method shows better orientation accuracy in both scenarios. Notably, in scenarios of partial observations, our model achieves +4\%, +22\%, and +16\% improvement on three orientation datasets as shown in Table \ref{tab1}. 

\subsubsection{Ablation Study}
\label{ablation}
For the accuracy contribution to the whole model, we conducted ablation experiments for the ViT backbone and foot joints. The results in Table \ref{tab2} show that ViT backbone and foot joint constraints contribute to +2.4\% and +0.3\% improvement, respectively, under the partial observation of the MEBOW dataset.

\subsubsection{Evaluation of Confidence}
\label{confidence}
The confidence in the range of $(0,1)$ indicates the reliability of orientation estimation,  and the baseline orientation output $\hat{\textbf{p}}$ represents a different probability of 72 class orientation. To evaluate the predicted confidence with the baseline method, we can use the orientation probability and confidence to classify the reliable and unreliable samples. 
Here, we utilize max recall @ 100\% precision as evaluation metrics for binary reliability classification accuracy. The max recall @ 100\% precision indicates the ability to find true orientation estimation without false positives. This ability is crucial for RPF because trusting a wrong estimation would result in dangerous RPF behavior.

\begin{table}[t]
    \centering
    % \captionsetup{justification=centering}
    \caption{Ablation study for the effectiveness of additional foot joint constraints and pre-trained ViT backbone}
    \label{tab2}
    \scalebox{0.9}{\begin{tabular}{c|c|c|ccc}
    \toprule
         \textbf{method} & \textbf{ViT} & \textbf{foot joints}  & \textbf{Acc (30°)} $\uparrow$ & \textbf{MAE (°)} $\downarrow$ \\ \midrule
         MEBOW & \XSolidBrush & \XSolidBrush &  90.0\% & 16.3 \\
         & \XSolidBrush & $\checkmark$ &   90.3\% (+0.3\%) & 16.2 (-0.1) \\
         &$\checkmark$ & \XSolidBrush & 92.4\% (+2.4\%) & 14.0 (-2.3) \\
         Ours &$\checkmark$ & $\checkmark$ & 93.4\% (+3.4\%) & 13.4 (-2.9) \\
         % Ours &$\checkmark$ & $\checkmark$ & 94.0\% (+4.0\%) & 12.0 (-4.3) \\
        \bottomrule
    \end{tabular}}
% \vspace*{-0.2in}
\end{table}

\begin{figure}[t]
\begin{minipage}{0.5\textwidth}
    \centering
    \begin{subfigure}{1.0\textwidth}
        \centering
        \includegraphics[width=\textwidth]{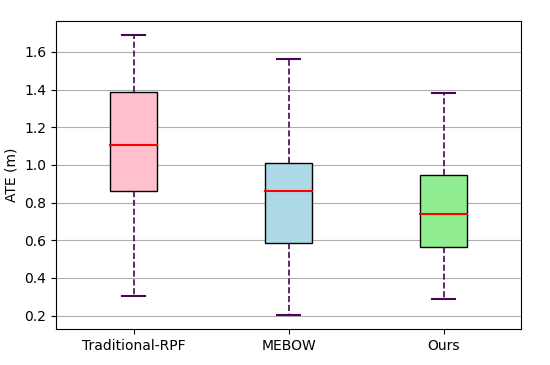}
        \caption{Backward RPF}
        \label{fig:image1}
    \vspace*{+0.1in}
    \end{subfigure}
    % \begin{subfigure}{0.49\textwidth}
    %     \centering
    %     \includegraphics[width=\textwidth]{compare_orientation_forward_baseline.png}
    %     \caption{MEBOW Orientation Estimation}
    %     \label{fig:image2}
    % \end{subfigure}
    % \\
    % \begin{subfigure}{0.49\textwidth}
    %     \centering
    %     \includegraphics[width=\textwidth]{Trajectory_exp/vit_forward_2.png}
    %     \caption{Ours forward}
    %     \label{fig:image3}
    % \end{subfigure}
    % \begin{subfigure}{0.49\textwidth}
    %     \centering
    %     \includegraphics[width=\textwidth]{Trajectory_exp/baseline_foward_2.png}
    %     \caption{MEBOW Forward}
    %     \label{fig:image4}
    % \end{subfigure}
    % \\
    % \begin{subfigure}{0.49\textwidth}
    %     \centering
    %     \includegraphics[width=\textwidth]{Trajectory_exp/comparison_backward.png}
    %     \caption{Backward ATE}
    %     \label{fig:image6}
    % \end{subfigure}
    \begin{subfigure}{1.0\textwidth}
        \centering
        \includegraphics[width=\textwidth]{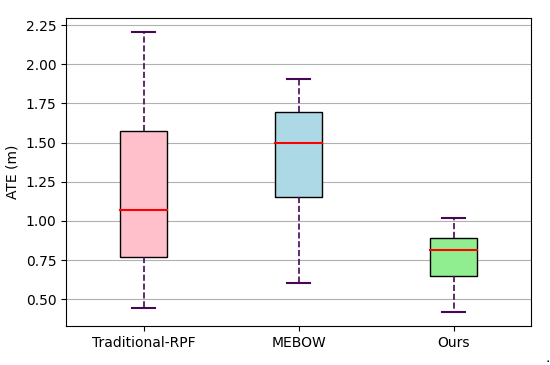}
        \caption{Forward RPF}
        \label{fig:image6}

    \end{subfigure}
    \end{minipage}
    % \captionsetup{justification=centering}

    \caption{Comparison of different RPF methods in real robot experiments. This comparison shows the absolute trajectory error (ATE) for different RPF methods, with the green box representing our method, the blue box representing MEBOW, and the red box representing the traditional RPF method that relies solely on human velocity for orientation. The lines above and below the dashed lines indicate the maximum and minimum values, while the red line represents the mean. In the two person-following scenarios, which include (a) Backward RPF task evaluation and (b) Forward RPF task evaluation, using PartHOE for orientation estimation demonstrates the best performance in person-following.} 
    \label{robotexp}
\vspace*{-0.3in}
\end{figure}

\begin{figure}[t]
\centering %表示居中
\includegraphics[width=1.0\linewidth]{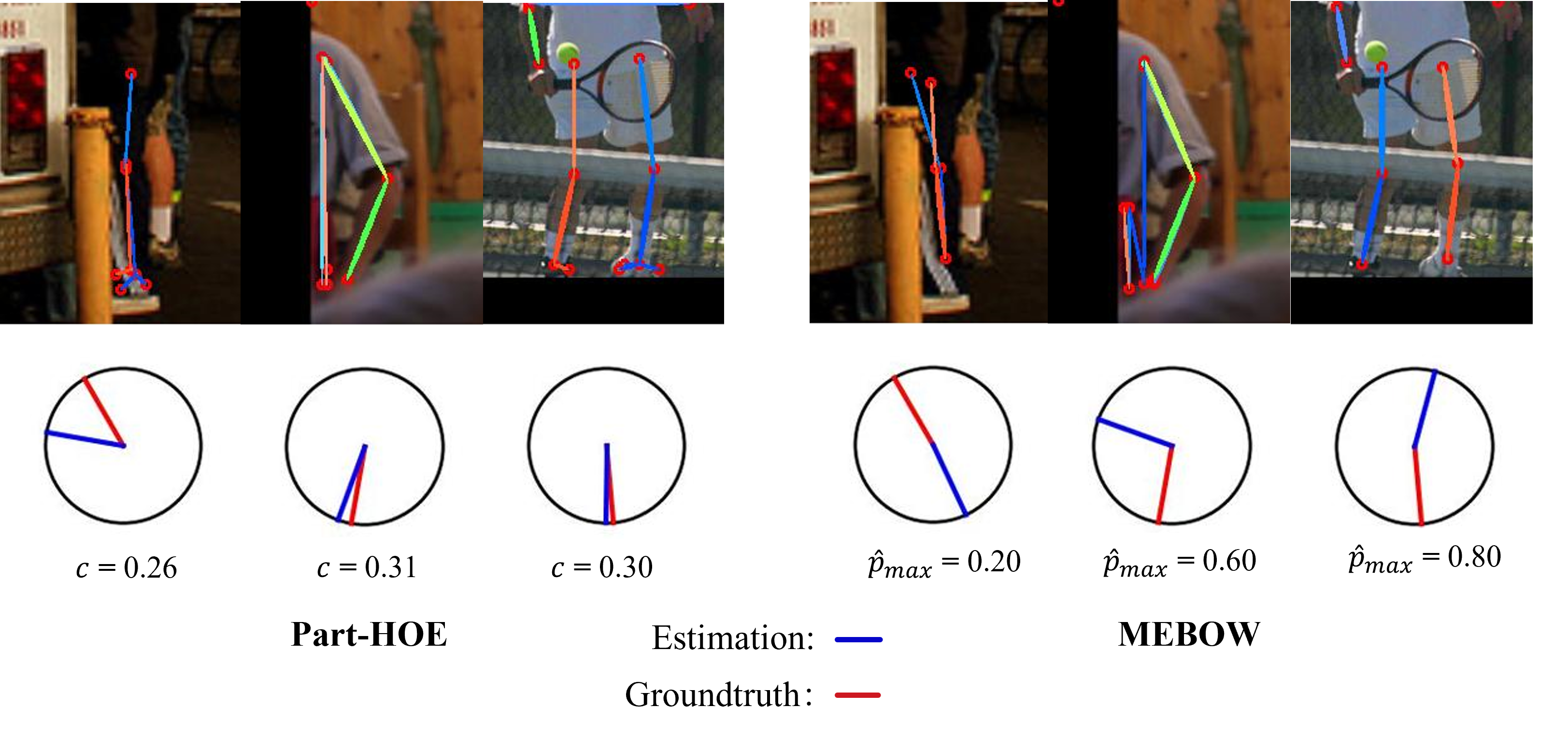}
% [height=4.5cm]表示高度
%[width=9.5cm]表示宽度
%{111.eps}表示eps格式的图片，名为111
\caption{MEBOW tends to output a high probability but incorrect prediction, as shown in the example above, while our Part-HOE provides a more reasonable confidence level. The variable $c$ represents the confidence output of PartHOE, while $\hat{p}_{\text{max}}$ denotes the maximum probability output of MEBOW.}
%图片的名称
\label{figure5}
%图片的标签，用于文章中的引用，注意到标签的数字与实际文章显示的数字可能不同
\end{figure}

Here, we set the samples with an orientation estimation error bigger than 20° as unreliable samples. We conduct experiments under a partial observation situation in the MEBOW dataset. The baseline MEBOW dropped from the very beginning 0.02 while we are able to keep 100\% precision till recall is 0.12 as shown in Fig. \ref{figure4},
% By setting the prediction error within $20^\circ$ as the reliable sample, otherwise unreliable. Thus, we are able to compare binary classification ability to reliable and unreliable samples using PR-curve shown in Fig. \ref{figure4} 

The baseline method makes wrong predictions with an unexpectedly high probability, and we can observe from Fig.~\ref{figure5} that MEBOW tends to give high probability output when there are enough visible joints and ignores the fact that only a few orientation cues can be observed (we scale the probability relative to its maximum value). Although such a scenario is hard for HOE because most orientation cues are occluded, our method can give reasonable confidence output and predict accurate orientation estimation.

\subsubsection{Real Robot Experiments}
\label{realrobot}
To further evaluate our model's accuracy and robustness, we integrate part-HOE into a robot person following (RPF) system as described in Sec.~\ref{RPFsystem}. There are two metrics used in real robot experiments: the absolute trajectory error (ATE) of the following trajectory and the orientation estimation accuracy after using confidence.
For the RPF task evaluation, we define the following forward and backward tasks, which correspond to the two cases where the robot accompanies the target person forward and backward at a fixed distance. Here, we set the following distance to 1 m.
By integrating our Part-HOE into the RPF system, we significantly reduce ATE by 0.65 m at the following forward task and reduce ATE by 0.31 m at the following backward task compared to the traditional RPF system, as shown in Fig.~\ref{robotexp} (b). We also compare MEBOW with the same experiment setting, and MEBOW achieves 1.23 m ATE at the following forward task and 0.82 m ATE at the following backward task. The MEBOW-based RPF system is better than the traditional RPF system but worse than our method. We can see that our following trajectory is closer to the ground truth compared to MEBOW, as shown in Fig.~\ref{robotexp} (a). 

\begin{figure}[t]
\centering %表示居中
\includegraphics[width=1.0\linewidth]{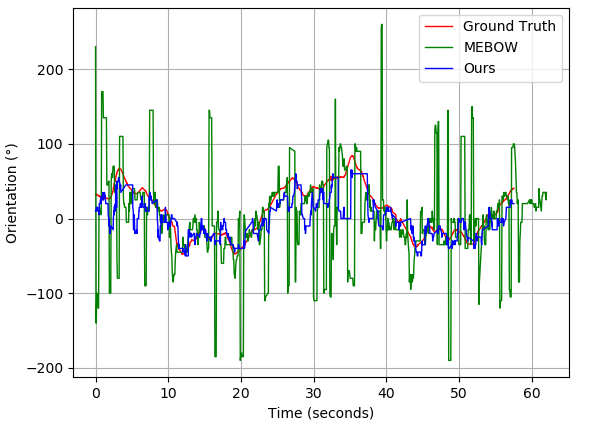}
% [height=4.5cm]表示高度
%[width=9.5cm]表示宽度
%{111.eps}表示eps格式的图片，名为111
\caption{Visualization of HOE during forward RPF task.}
%图片的名称
\label{orientationcurve}
%图片的标签，用于文章中的引用，注意到标签的数字与实际文章显示的数字可能不同
\end{figure}

We further evaluate the confidence output of our orientation estimation model. The unreliable samples are filtered out during the following process, and we just directly set the orientation at the current frame to the most confident estimation among the last 5 frames since the frequency of our model is about 25 FPS. The orientation filtered by our confidence achieves higher accuracy compared to the orientation filtered by MEBOW probability and shows slight variation as shown in Fig.~\ref{orientationcurve}.

\section{CONCLUSION} \label{sec:conclusion}
In this paper, we show that enhanced joint detection, particularly with a pre-trained ViT model and additional foot joints, significantly improves orientation accuracy under occlusion, with gains of up to 22\% on the Human3.6M dataset and 16\% on our custom dataset. The proposed self-supervised method for confidence estimation offers more reliable filtering of uncertain samples compared to traditional approaches like MEBOW. Additionally, the proposed Part-HOE method demonstrates superiority in real robot applications, i.e., robot person following. However, Part-HOE has yet to fully utilize temporal information and still struggles with effectively filtering out unreliable orientation estimates. We aim to address these issues in our future work.
% \addtolength{\textheight}{-12cm}   % This command serves to balance the column lengths
                                  % on the last page of the document manually. It shortens
                                  % the textheight of the last page by a suitable amount.
                                  % This command does not take effect until the next page
                                  % so it should come on the page before the last. Make
                                  % sure that you do not shorten the textheight too much.

%%%%%%%%%%%%%%%%%%%%%%%%%%%%%%%%%%%%%%%%%%%%%%%%%%%%%%%%%%%%%%%%%%%%%%%%%%%%%%%%

%%%%%%%%%%%%%%%%%%%%%%%%%%%%%%%%%%%%%%%%%%%%%%%%%%%%%%%%%%%%%%%%%%%%%%%%%%%%%%%%

%%%%%%%%%%%%%%%%%%%%%%%%%%%%%%%%%%%%%%%%%%%%%%%%%%%%%%%%%%%%%%%%%%%%%%%%%%%%%%%%
% \section*{APPENDIX}

\bibliographystyle{IEEEtran}
\bibliography{ref}

\begin{thebibliography}{10}
\providecommand{\url}[1]{#1}
\csname url@rmstyle\endcsname
\providecommand{\newblock}{\relax}
\providecommand{\bibinfo}[2]{#2}
\providecommand\BIBentrySTDinterwordspacing{\spaceskip=0pt\relax}
\providecommand\BIBentryALTinterwordstretchfactor{4}
\providecommand\BIBentryALTinterwordspacing{\spaceskip=\fontdimen2\font plus
\BIBentryALTinterwordstretchfactor\fontdimen3\font minus
  \fontdimen4\font\relax}
\providecommand\BIBforeignlanguage[2]{{%
\expandafter\ifx\csname l@#1\endcsname\relax
\typeout{** WARNING: IEEEtran.bst: No hyphenation pattern has been}%
\typeout{** loaded for the language `#1'. Using the pattern for}%
\typeout{** the default language instead.}%
\else
\language=\csname l@#1\endcsname
\fi
#2}}

\bibitem{islam2019person}
M.~J. Islam, J.~Hong, and J.~Sattar, ``Person-following by autonomous robots: A
  categorical overview,'' \emph{The International Journal of Robotics
  Research}, vol.~38, no.~14, pp. 1581--1618, 2019.

\bibitem{12walkingaid}
K.~Wakita, J.~Huang, P.~Di, K.~Sekiyama, and T.~Fukuda,
  ``Human-walking-intention-based motion control of an omnidirectional-type
  cane robot,'' \emph{IEEE/ASME Transactions on Mechatronics}, vol.~18, no.~1,
  pp. 285--296, 2013.

\bibitem{20drones}
A.~Ashtari, S.~Stev{\v{s}}i{\'c}, T.~N{\"a}geli, J.-C. Bazin, and O.~Hilliges,
  ``Capturing subjective first-person view shots with drones for automated
  cinematography,'' \emph{ACM Transactions on Graphics (TOG)}, vol.~39, no.~5,
  pp. 1--14, 2020.

\bibitem{18Nikdel}
P.~Nikdel, R.~Shrestha, and R.~Vaughan, ``The hands-free push-cart: Autonomous
  following in front by predicting user trajectory around obstacles,'' in
  \emph{2018 IEEE International Conference on Robotics and Automation (ICRA)},
  2018, pp. 4548--4554.

\bibitem{leigh2015person}
A.~Leigh, J.~Pineau, N.~Olmedo, and H.~Zhang, ``Person tracking and following
  with 2d laser scanners,'' in \emph{2015 IEEE International Conference on
  robotics and automation (ICRA)}.\hskip 1em plus 0.5em minus 0.4em\relax IEEE,
  2015, pp. 726--733.

\bibitem{19kpt}
D.~Yu, H.~Xiong, Q.~Xu, J.~Wang, and K.~Li, ``Continuous pedestrian orientation
  estimation using human keypoints,'' in \emph{2019 IEEE International
  Symposium on Circuits and Systems (ISCAS)}, 2019, pp. 1--5.

\bibitem{21monoloco}
L.~Bertoni, S.~Kreiss, and A.~Alahi, ``Perceiving humans: from monocular 3d
  localization to social distancing,'' \emph{IEEE Transactions on Intelligent
  Transportation Systems}, vol.~23, no.~7, pp. 7401--7418, 2021.

\bibitem{mebow}
C.~Wu, Y.~Chen, J.~Luo, C.-C. Su, A.~Dawane, B.~Hanzra, Z.~Deng, B.~Liu, J.~Z.
  Wang, and C.-h. Kuo, ``Mebow: Monocular estimation of body orientation in the
  wild,'' in \emph{Proceedings of the IEEE/CVF Conference on Computer Vision
  and Pattern Recognition (CVPR)}, June 2020.

\bibitem{hrnet}
K.~Sun, B.~Xiao, D.~Liu, and J.~Wang, ``Deep high-resolution representation
  learning for human pose estimation,'' in \emph{Proceedings of the IEEE/CVF
  conference on computer vision and pattern recognition}, 2019, pp. 5693--5703.

\bibitem{21udp}
H.~Liu, F.~Liu, X.~Fan, and D.~Huang, ``Polarized self-attention: Towards
  high-quality pixel-wise regression,'' \emph{arXiv preprint arXiv:2107.00782},
  2021.

\bibitem{wang2023unbiased}
C.~Wang, Y.~Zhou, F.~Zhang, and P.~Mok, ``Unbiased feature position alignment
  for human pose estimation,'' \emph{Neurocomputing}, vol. 537, pp. 152--163,
  2023.

\bibitem{coco}
T.-Y. Lin, M.~Maire, S.~Belongie, J.~Hays, P.~Perona, D.~Ramanan,
  P.~Doll{\'a}r, and C.~L. Zitnick, ``Microsoft coco: Common objects in
  context,'' in \emph{Computer Vision--ECCV 2014: 13th European Conference,
  Zurich, Switzerland, September 6-12, 2014, Proceedings, Part V 13}.\hskip 1em
  plus 0.5em minus 0.4em\relax Springer, 2014, pp. 740--755.

\bibitem{23vitpose}
Y.~Xu, J.~Zhang, Q.~Zhang, and D.~Tao, ``Vitpose++: Vision transformer for
  generic body pose estimation,'' \emph{IEEE Transactions on Pattern Analysis
  and Machine Intelligence}, 2023.

\bibitem{vit}
A.~Dosovitskiy, L.~Beyer, A.~Kolesnikov, D.~Weissenborn, X.~Zhai,
  T.~Unterthiner, M.~Dehghani, M.~Minderer, G.~Heigold, S.~Gelly,
  \emph{et~al.}, ``An image is worth 16x16 words: Transformers for image
  recognition at scale,'' \emph{arXiv preprint arXiv:2010.11929}, 2020.

\bibitem{mp2d}
M.~Andriluka, L.~Pishchulin, P.~Gehler, and B.~Schiele, ``2d human pose
  estimation: New benchmark and state of the art analysis,'' in
  \emph{Proceedings of the IEEE Conference on computer Vision and Pattern
  Recognition}, 2014, pp. 3686--3693.

\bibitem{AIC}
J.~Wu, H.~Zheng, B.~Zhao, Y.~Li, B.~Yan, R.~Liang, W.~Wang, S.~Zhou, G.~Lin,
  Y.~Fu, \emph{et~al.}, ``Ai challenger: A large-scale dataset for going deeper
  in image understanding,'' \emph{arXiv preprint arXiv:1711.06475}, 2017.

\bibitem{cocowhole}
S.~Jin, L.~Xu, J.~Xu, C.~Wang, W.~Liu, C.~Qian, W.~Ouyang, and P.~Luo,
  ``Whole-body human pose estimation in the wild,'' in \emph{Computer
  Vision--ECCV 2020: 16th European Conference, Glasgow, UK, August 23--28,
  2020, Proceedings, Part IX 16}.\hskip 1em plus 0.5em minus 0.4em\relax
  Springer, 2020, pp. 196--214.

\bibitem{10tud}
M.~Andriluka, S.~Roth, and B.~Schiele, ``Monocular 3d pose estimation and
  tracking by detection,'' in \emph{2010 IEEE Computer Society Conference on
  Computer Vision and Pattern Recognition}, 2010, pp. 623--630.

\bibitem{deep-orientation}
B.~Lewandowski, D.~Seichter, T.~Wengefeld, L.~Pfennig, H.~Drumm, and H.-M.
  Gross, ``Deep orientation: Fast and robust upper body orientation estimation
  for mobile robotic applications,'' in \emph{2019 IEEE/RSJ International
  Conference on Intelligent Robots and Systems (IROS)}, 2019, pp. 441--448.

\bibitem{18gaze}
T.~Fischer, H.~J. Chang, and Y.~Demiris, ``Rt-gene: Real-time eye gaze
  estimation in natural environments,'' in \emph{Proceedings of the European
  conference on computer vision (ECCV)}, 2018, pp. 334--352.

\bibitem{21pifpaf}
S.~Kreiss, L.~Bertoni, and A.~Alahi, ``Openpifpaf: Composite fields for
  semantic keypoint detection and spatio-temporal association,'' \emph{IEEE
  Transactions on Intelligent Transportation Systems}, vol.~23, no.~8, pp.
  13\,498--13\,511, 2021.

\bibitem{resnet}
K.~He, X.~Zhang, S.~Ren, and J.~Sun, ``Deep residual learning for image
  recognition,'' in \emph{Proceedings of the IEEE conference on computer vision
  and pattern recognition}, 2016, pp. 770--778.

\bibitem{learnconfidence}
T.~DeVries and G.~W. Taylor, ``Learning confidence for out-of-distribution
  detection in neural networks,'' \emph{arXiv preprint arXiv:1802.04865}, 2018.

\bibitem{08walkingaid}
J.~Huang, P.~Di, T.~Fukuda, and T.~Matsuno, ``Motion control of
  omni-directional type cane robot based on human intention,'' in \emph{2008
  IEEE/RSJ International Conference on Intelligent Robots and Systems}, 2008,
  pp. 273--278.

\bibitem{22walkingaid}
Q.~Yan, J.~Huang, Z.~Yang, Y.~Hasegawa, and T.~Fukuda, ``Human-following
  control of cane-type walking-aid robot within fixed relative posture,''
  \emph{IEEE/ASME Transactions on Mechatronics}, vol.~27, no.~1, pp. 537--548,
  2022.

\bibitem{ye2023icra}
H.~Ye, J.~Zhao, Y.~Pan, W.~Chen, L.~He, and H.~Zhang, ``Robot person following
  under partial occlusion,'' in \emph{2023 IEEE International Conference on
  Robotics and Automation (ICRA)}, 2023, pp. 7591--7597.

\bibitem{ye2023person}
H.~Ye, J.~Zhao, Y.~Zhan, W.~Chen, L.~He, and H.~Zhang, ``Person
  re-identification for robot person following with online continual
  learning,'' \emph{IEEE Robotics and Automation Letters}, pp. 1--8, 2024.

\bibitem{human36m}
C.~Ionescu, D.~Papava, V.~Olaru, and C.~Sminchisescu, ``Human3. 6m: Large scale
  datasets and predictive methods for 3d human sensing in natural
  environments,'' \emph{IEEE transactions on pattern analysis and machine
  intelligence}, vol.~36, no.~7, pp. 1325--1339, 2013.

\bibitem{ge2021yolox}
Z.~Ge, S.~Liu, F.~Wang, Z.~Li, and J.~Sun, ``Yolox: Exceeding yolo series in
  2021,'' \emph{arXiv preprint arXiv:2107.08430}, 2021.

\end{thebibliography}
\end{CJK}
\end{document}